\documentclass{article}

\usepackage{neurips_2025}

\usepackage[utf8]{inputenc} 
\usepackage[T1]{fontenc}    
\usepackage{hyperref}       
\usepackage{url}            
\usepackage{booktabs}       
\usepackage{amsfonts}       
\usepackage{nicefrac}       
\usepackage{microtype}      
\usepackage{xcolor}         
\usepackage{graphicx}
\usepackage{algorithm}
\usepackage{algpseudocode}
\usepackage{amsmath}

\title{Balancing Computation Load and Representation Expressivity in Parallel Hybrid Neural Networks}

\author{%
  Mohammad Mahdi Moradi\\
  Department of Computer Science, Concordia University\\
  Ascend Team, Huawei Technologies\\
  \texttt{mohammad.mahdi.moradi@h-partners.com} \\
  \And
  Walid Ahmed\\
  Ascend Team \\
  Toronto Research Center \\
  Huawei Technologies \\
  \AND
  Shuangyue Wen\\
  Ascend Team \\
  Toronto Research Center \\
  Huawei Technologies \\
  \AND
  Sudhir  Mudur \\
  Department of Computer Science \\
  Concordia University \\
  \texttt{mudur@cs.concordia.ca} \\
  \And
  Weiwei Zhang\\
  Ascend Team \\
  Toronto Research Center \\
  Huawei Technologies \\
  \And
  Yang Liu\\
  Ascend Team \\
  Toronto Research Center \\
  Huawei Technologies \\  
}

\begin{document}

\maketitle

\begin{abstract}\label{sec:abstract}
 Attention and State-Space Models (SSMs) when combined in a hybrid network in sequence or in parallel provide complementary strengths. In a hybrid sequential pipeline they alternate between applying a transformer to the input and then feeding its output into a SSM. This results in idle periods in the individual components increasing end‐to‐end latency and lowering throughput caps. In the parallel hybrid architecture, the transformer operates independently in parallel with the SSM, and these pairs are cascaded,  with output from one pair forming the input to the next. Two issues are (i) creating an expressive knowledge representation with the inherently divergent outputs from these separate branches, and (ii) load balancing the computation between these parallel branches, while maintaining representation fidelity. In this work we present FlowHN, a novel parallel hybrid network architecture that accommodates various strategies for load balancing, achieved through appropriate distribution of input tokens between the two branches. Two innovative differentiating factors in FlowHN include a FLOP aware dynamic token split between the attention and SSM branches yielding efficient balance in compute load, and secondly, a method to fuse the highly divergent outputs from individual branches for enhancing representation expressivity. Together they enable much better token processing speeds, avoid bottlenecks, and at the same time yield significantly improved accuracy as compared to other competing works. We conduct comprehensive experiments on autoregressive language modeling for models with 135M, 350M, and 1B parameters. FlowHN outperforms sequential hybrid models and its parallel counterpart, achieving up to 4× higher Tokens per Second (TPS) and 2× better Model FLOPs Utilization (MFU).
\end{abstract}

\section{Introduction}\label{sec:Introduction}
Transformer-based language models have become the foundation of modern Natural Language Processing (NLP) thanks to their ability to learn rich contextual representations through their self-attention mechanism for capturing fine-grained token interactions. While long-range relationships can also be captured, the quadratic cost of self-attention, requiring pairwise token interactions across the entire context grows prohibitively expensive in both computation and memory as sequence lengths increase (\cite{zeng2025zeta}). To address the limitations of traditional attention mechanisms in handling long sequences, researchers have proposed various alternatives, including Linear Attention (\cite{katharopoulos2020transformers}), Gated Linear Attention (\cite{yang2023gated}), and more recently, State Space Models (SSMs) (\cite{fu2022hungry, gu2023mamba}). SSMs offer constant-time recurrence and hardware-friendly implementations, making them efficient for long-range dependencies. However, while SSMs excel in modeling long range interactions with low computational overhead, they often under-perform compared to Transformers on tasks requiring fine-grained token interactions.

\begin{figure*}[t]
  \centering
   \includegraphics[width=\textwidth]{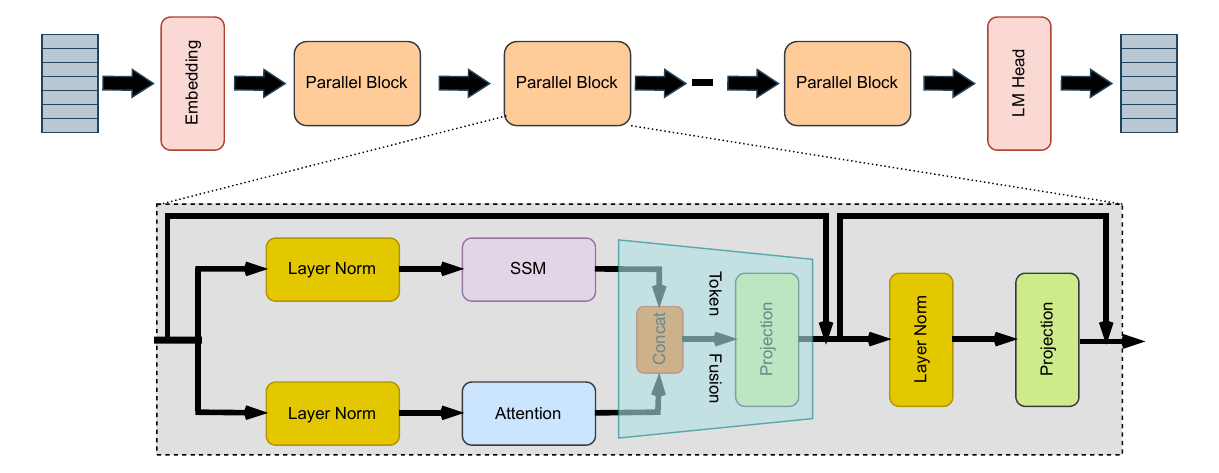}
   \caption{The overall architecture of our proposed parallel hybrid model, FlowHN}
   \label{fig:PHMarchitecture}
\end{figure*}

Hybrid networks have thus been proposed to combine the strengths of both, attention and SSM mechanisms to capture both short and long range dependencies. The two can be operative in sequence or in parallel. Our work presented in this paper is in the parallel category. Hymba (\cite{dong2024hymba}) is the only other parallel hybrid model that we are aware of with the same objective. The novelty in our proposed architecture lies in two important aspects. The first is a method for dynamic load balancing between the attention and SSM sub-blocks through appropriate distribution of input tokens with low information loss.  The second novelty is a method to integrate the inherently divergent information coming independently from the two branches, reduce the information loss caused by independent parallel processing and yet capture knowledge effectively. These methods will be described in detail in later sections. We call this hybrid network as FlowHN. 


Our main contributions are:
\begin{itemize}
    \item A novel parallel block (Figure \ref{fig:PHMarchitecture}) that processes each input through two concurrent branches, an attention branch for capturing global context and a SSM branch for memory-efficient, long-range dependency capture. The inherently divergent outputs from these concurrent branches are fused and projected to obtain a more expressive combined representation that forms the input to the next parallel block. 

    \item  We present three token distribution strategies, including a dynamic FLOP-aware circulating split algorithm that adaptively distributes tokens between the two branches within each parallel block, thus balancing workload between branches and reducing total compute time with only slight information loss as compared to no splitting.   

    \item Through extensive experiments on standard text benchmarks and across three model scales (135M, 350M, and 1B parameters), we demonstrate that our architecture consistently surpasses both its parallel counterpart and leading sequential architectures in accuracy and computational efficiency.
\end{itemize}

\section{Related Works}\label{sec:Related Works}
Large-scale autoregressive language models have demonstrated remarkable sequence processing capabilities, ranging from zero and few shot learning to long-range reasoning by pretraining on massive text corpora. However, their reliance on quadratic self-attention, which scales as $O(N^2)$ in sequence length, imposes steep compute and memory demands that hinder processing of long contexts  (\cite{katharopoulos2020transformers}). This challenge has spurred two complementary lines of research to reduce/eliminate self-attention compute costs, subquadratic attention approximations and state-space models, each with their individual strengths.

\textbf{Subquadratic Sequence Modeling. } Early work on reducing the quadratic bottleneck reformulated self-attention as a linear kernelized dot-product, enabling $O(N)$ complexity via associativity of matrix products (e.g., Linear Attentions,  
\cite{katharopoulos2020transformers}). Concurrently, RNN-inspired architectures such as RWKV blend parallelizable training with recurrent inference to achieve subquadratic scaling without softmax attention (\cite{peng2023rwkv}). These approaches maintain competitive performance on long-sequence tasks while dramatically cutting FLOPs and memory usage. Yet they may struggle with some problems which full attention transformers are good at, such as content based reasoning. Although subquadratic sequence models dramatically reduce FLOPs and memory, they still struggle to preserve high-fidelity and long-range dependencies (\cite{han2024demystify}) and suffer from representation collapse due to the non-injective nature of linearized attention kernels, which can assign identical weights to different queries (\cite{han2024bridging}).
 
\textbf{State-Space Models. } Recently attention-free sequence modeling has centered on SSM architectures, which efficiently capture long-range dependencies without the quadratic cost of self-attention. The Structured State Space Sequence (S4) model (\cite{gu2021efficiently}) introduced a novel parameterization of the linear SSM kernel that yields fast convolutional approximations and achieves competitive performance on long sequences. Building on S4, the H3 architecture (\cite{fu2022hungry}) interleaves shift and diagonal SSM matrices with multiplicative input projections, substantially narrowing the expressivity gap between pure SSMs and Transformer layers. Mamba (\cite{gu2023mamba}) further unifies these selective SSM blocks with standard MLP layers, employing a selective-scan algorithm and hardware-aware optimizations (parallel scan, kernel fusion, and recomputation) to achieve true linear-time scaling and up to 5x higher TPS (Tokens/sec) than comparable Transformer-based models across language, audio, and genomics benchmarks. More recently, Mamba-2 (\cite{dao2024transformers}) leverages structured state-space duality (SSD) to simplify parameter generation, producing SSM coefficients at block inception rather than as input-dependent functions, yielding 2–8x speedups over the original Mamba while delivering superior modeling quality relative to Transformer baselines. 

\textbf{Sequential Hybrid Models. } Studies have highlighted that purely linear SSMs exhibit limited retrieval capabilities and thus benefit from integrating attention mechanisms to restore precise memory access (\cite{jelassi2024repeat, abdin2024phi, mohtashami2023landmark}). MambaFormer (\cite{park2024can}) extends this hybridization by replacing MLP blocks in standard Transformers with selective Mamba SSM blocks, thereby leveraging recurrent state updates to enhance long-range in-context learning and retrieval tasks. Jamba (\cite{lieber2024jamba, team2024jamba}) adopts a mixture-of-experts (MoE) design, interleaving Transformer, Mamba, and MoE layers to achieve high TPS, reduced memory footprint, and robust performance on contexts up to 256K tokens. SAMBA (\cite{ren2024samba}) presents a lightweight, layer-wise fusion of selective SSM (Mamba) with the Sliding Window Attention technique, compressing distant context into recurrent hidden states while preserving fine-grained local attention, thus enabling efficient modeling of effectively unlimited sequences. Despite these advances, existing sequential hybrids are unable to avoid the bottlenecks of sequential stacking and thus do not efficiently exploit both attention and recurrence.

\textbf{Parallel Hybrid Models. } In \cite{dong2024hymba} Hymba, a family of small language models featuring a hybrid‐head parallel architecture is introduced. To further bolster contextual recall, Hymba makes use of learnable meta tokens prepended to input sequences, which serve as a shared, dynamic cache across attention and SSM branches, thereby concentrating model capacity on salient information.  The hybrid‐head module is also optimized via a mix of global and partial sliding‐window local attention, combined with cross‐layer key–value cache sharing to sharply reduce memory overhead without compromising recall fidelity. However, despite these enhancements, Hymba faces computational efficiency challenges: the two branches process inputs at different speeds, causing the faster branch to idle while waiting for the slower one, ultimately leading to TPS bottlenecks. Another disadvantage is that the use of learnable meta tokens when concatenated to the input sequence increases the sequence length, which permeates through all stages of the pipeline and thus introduces additional computation/memory overheads. 

In comparison, our FlowHN architecture balances the computational work load across attention and SSMs more evenly, does not require learnable meta tokens, so processing of extraneous tokens is not involved, and importantly, fuses and projects the outputs from independent branches to yield more expressive knowledge representation, as can be seen from our experimental comparisons. 

\begin{figure*}[h]
  \centering
   \includegraphics[width=\textwidth]{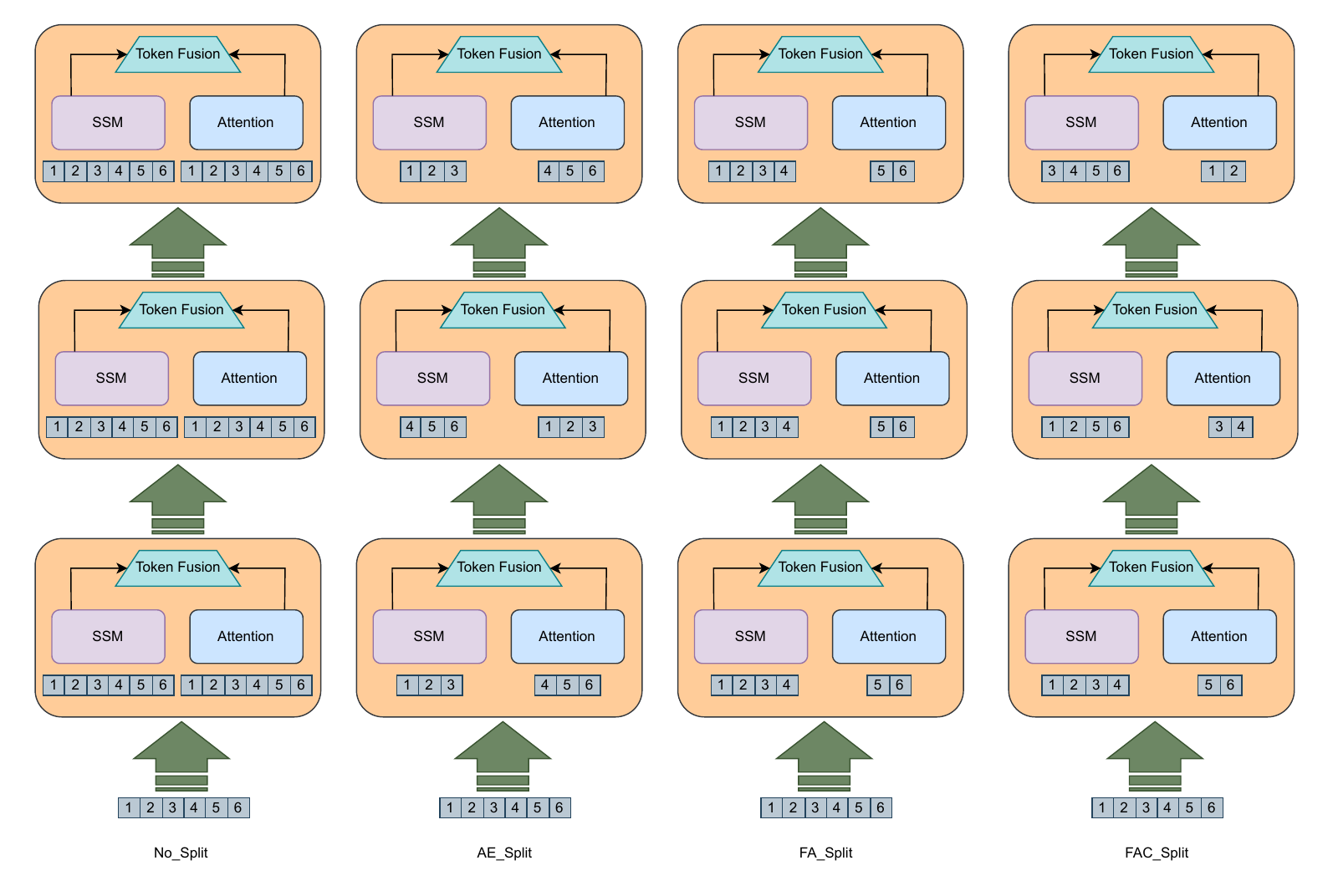}
   \caption{Illustrative example of our four token-splitting strategies in action, showing how input tokens are partitioned and processed under each method.}
   \label{fig:PHM_modes}
\end{figure*}

\section{Methodology}\label{sec:Methodology}
Figure \ref{fig:PHMarchitecture} illustrates the overall architecture of our parallel hybrid network, FlowHN. It is composed of a cascade of parallel blocks, each of which contains two branches operating side by side: a SSM branch and an Attention branch. To allow the model to adaptively balance computation load between these two independent processing pathways, we introduce a dynamic token-splitting mechanism that distributes the input tokens between the parallel branches. While there is some information loss caused by the fact that components process the sequence not as a whole but in disjoint subsets, our split strategies are geared towards mitigating such loss. We explore three token splitting and allocation strategies. Each successor is an improvement over its predecessor in terms of load balancing between the two branches, as summarized in Figure \ref{fig:PHM_modes}. In the sections that follow, we describe in detail how each strategy partitions and routes tokens, how the branches process their respective input subsets, and how their outputs are subsequently fused and projected to improve the expressivity and yield the final representation from the parallel block. 

\subsection{No\_Split Token Allocation Strategy}\label{sec:NoSplit Token Allocation Strategy}
In this strategy called No\_Split (Figure \ref{fig:PHM_modes}), the entire input token sequence is concurrently routed to both the SSM and Attention branches without any split. To reconcile the inherently divergent output formats of these pathways, we integrate a Token Fusion sub-module (depicted in blue in Figure \ref{fig:PHMarchitecture}) that serves as an information aggregator. Concretely, the dual-branch outputs are concatenated (not averaged as in Hymba (\cite{dong2024hymba})) and then passed through a linear projection, yielding a fused representation of the prescribed dimensionality. This fusion operation produces a semantically enriched embedding that retains complementary features from both branches while promoting feature interplay across modalities. Consequently, the network learns to optimally integrate these heterogeneous features, thereby enhancing the expressiveness and robustness of the resulting representation. As can be seen from our comparison results in Table ~\ref{table:1}, our fusion operation yields comparable (at times better) representation fidelity to that obtained through the use of meta-tokens in Hymba, and at much lower computational costs.

\subsection{Alternating\_Equal\_Split Token Allocation Strategy }\label{sec:AlternatingEqualSplit Token Allocation Strategy}
In this strategy called AE\_Split, (Figure \ref{fig:PHM_modes}), the input token sequence is evenly partitioned between the SSM and Attention branches to halve the computational burden of each branch and accelerate the overall processing. Because each branch only sees half of the input in any given block, potentially relevant contextual information residing in the complementary half would be missed. To address this, we employ an alternating‐half scheme whereby, in the subsequent block, each branch processes the opposite half of the sequence that it skipped previously. This “Alternative” scheduling thus ensures that over two consecutive block, both branches attend to all tokens, mitigating information loss while retaining per-block speed gains. Nevertheless, the faster branch must still idle until its slower counterpart completes its computation before the fused representation can be passed onward. This causes a synchronization bottleneck that limits the computational efficiency of the model.

\begin{algorithm}[t]
\caption{FLOP-Aware Circulating Split (FAC\_Split) strategy}
\begin{algorithmic}[1]
\State \textbf{Input:} T = [$t_1$, $t_2$, ..., $t_L$]\label{algorithm:1}
\State \textbf{Output:} Processed results from SSM and Attention branches
\State \textbf{L:} Number of tokens in the input sequence
\State \textbf{T:} Sequence of input tokens
\State \textbf{F\_a:} Number of Attention FLOPs
\State \textbf{F\_s:} Number of SSM FLOPs
\State \textbf{block\_index} Number specifying block index
\State $block\_size \gets \frac{L}{\frac{F_s}{F_a}+1}$ \Comment{SSM block processes $\frac{L}{\frac{F_s}{F_a}+1}$ tokens}
\State $block\_index \gets 0$

\For{$i = 0$ to N-1} \Comment{N is Number of Parallel Blocks}
    \State $start\_index \gets (block\_index) \times block\_size$
    \State $end\_index \gets (block\_index+1) \times block\_size$
    \If{$end\_index < L$}
        \State $T_{SSM} \gets T[start\_index:end\_index)$ \Comment{block\_size tokens to SSM}
        \State $T_{Attention} \gets T[:start\_index \& end\_index:]$ \Comment{Rest of the tokens to Attention}
        
        \State $SSM_{Output} \gets SSM(T_{SSM})$
        \State $Attention_{Output} \gets Attention(T_{Attention})$
        
        \State \textbf{Merge} $SSM_{Output}$ and $Attention_{Output}$ \Comment{Merge results of SSM and Attention}
        \State $block\_index \gets block\_index + 1$
    \Else
        \State $block\_index \gets 0$
    \EndIf
\EndFor
\end{algorithmic}
\end{algorithm}

\subsection{Flop\_Aware\_Split Token Allocation Strategy}\label{sec:FlopAwareSplit Token Allocation Strategy}
This token allocation strategy called FA\_Split introduces a FLOP-aware token-splitting mechanism to mitigate the synchronization bottleneck inherent in alternating branch design of the earlier strategy. Tokens are allocated to the SSM and Attention branches according to their respective FLOPs per token requirement. A smaller subset of the input sequence is dynamically assigned to the branch with higher FLOPs per token requirement, and the remaining larger subset to the other branch.  This helps equitably divide the computational workload. FA\_Split synchronizes branch completion times and thereby approximates maximally parallel execution, resulting in improved TPS and memory efficiency. Despite these advantages, this strategy, while preserving the original input order, inevitably processes disjoint subsets of tokens in each branch, thus not facilitating cross-branch information sharing, so essential for increasing representation expressivity. As shown in Figure \ref{fig:PHM_modes}, this hard split can lead to missing contextual cues from other branches, which can limit the model's ability to capture global dependencies.

\subsection{FLOP\_Aware\_Circulating\_Split Token Allocation Strategy}\label{sec:Flop_Aware_Circulating_Split Token Allocation Strategy}
With the goal of overcoming the limitations of previous modes, we propose a FLOP-Aware Circulating Split strategy (FAC\_Split), that dynamically distributes tokens between the SSM and the Attention based on their relative computational costs. As illustrated in Figure \ref{fig:PHM_modes}, we first measure the FLOPs of each branch (SSM versus Attention) and then allocate tokens inversely proportional to those costs: the slower, higher‐FLOP cost branch receives fewer tokens, while the faster, lower‐FLOP cost branch receives more. For example in Figure \ref{fig:PHM_modes}, when the Attention’s FLOPs costs are twice those of the SSM, it processes only half that of the other branch in each parallel block. For example, in the first block, tokens 1–4 go to the SSM and tokens 5–6 to the Attention; in the next block, tokens 1–2 and 5–6 go to the SSM (with tokens 3–4 going to the Attention); and in the third block, tokens 1–2 go to the Attention and tokens 3–6 to the SSM. This cyclic reassignment ensures that, over the sequence of parallel blocks, both branches very likely see every token at least once. By balancing per‐branch workloads, this token splitting mode achieves higher TPS and better memory efficiency than static splits, and by rotating token assignments, it promotes richer, more diverse feature learning without introducing synchronization bottlenecks. Algorithm \ref{algorithm:1} details the token‐feeding procedure of FAC\_Split strategy.


\section{Experiments}\label{sec:Experiments}
\subsection{Settings}\label{sec:Settings}
\textbf{Datasets. } All models are trained on SlimPajama-6B, a 6 billion-token English subset sampled from the first chunk of the deduplicated SlimPajama-627B \cite{cerebras2023slimpajama} corpus that preserves its original domain mix (54.1\% CommonCrawl, 28.7\% C4, 4.2\% GitHub code, 3.7\% books, 3.4\% arXiv, 3.1\% Wikipedia, 2.8\% StackExchange), using the publicly available Cerebras preprocessing scripts.

\textbf{Baseline Models. } To assess the effectiveness of FlowHN, we benchmarked it against several state-of-the-art baselines—including LLaMA, Mamba-1, Mamba-2, Jamba, Sequential Hybrid Model (SHM), and Hymba—across three model scales: 135M, 350M, and 1B parameters. SHM serves as the sequential counterpart to our proposed FlowHN, interleaving LayerNorm + SSM + LayerNorm + MLP with LayerNorm + Attention + LayerNorm + MLP in sequence.

\subsection{Implementation details}\label{sec:Implementation details}
All experiments were conducted on a single NVIDIA Tesla V100 PCIe GPU with 32GB of memory. All models were trained using a single GPU. The average training times for models with 135M, 350M, and 1B parameters were around 17 hours, 3 days, and 11 days, respectively. Due to limited resources, we were unable to train our proposed model, FlowHN, on massive datasets like those used for large-scale models such as LLaMA, Mamba, Jamba, and Hymba. For a fair and rigorous comparison among all state of the art models, we trained all models, FlowHN and the baselines, from scratch under identical experimental conditions, even if these conditions may not be individually optimal, and not for FlowHN either. Specifically, each configuration was trained using exactly 1,024 tokens. For the 135M and 350M parameter models, we used a batch size of 16 with gradient accumulation over 8 steps, an initial learning rate of $3 \times 10^{-4}$, weight decay of 0.1, and AdamW with $\beta_1 = 0.9$ and $\beta_2=0.95$. We applied a cosine learning‐rate schedule with 10\% linear warmup and trained for a total of 1B tokens. For the 1B parameter models, we reduced the batch size to $4$ and increased gradient accumulation to 16 to accommodate memory constraints, set the learning rate to $1 \times 10^{-4}$, maintained the $0.1$ weight decay, and used AdamW with $\beta_1 = 0.9$ and $\beta_2 = 0.999$ under the same cosine schedule and 10\% warmup, training across 2B tokens.

\subsection{Results and Discussion}\label{sec:Results and Discussion}
Table \ref{table:1} presents the performance of FlowHN with the four token allocation strategies and three model sizes (135M, 350M, and 1B parameters), compared against Llama, Mamba-1, Mamba-2, Jamba, the  and the parallel hybrid model Hymba. 
This table also presents a comparison of the evaluated models in terms of Tokens per Second (TPS), Model FLOPs Utilization (MFU), and accuracy. MFU \cite{chowdhery2023palm} quantifies the efficiency with which the available computational resources (e.g., GPUs or TPUs) are utilized during both the forward and backward passes to perform floating-point operations—an essential component of neural network computations. A higher MFU indicates better utilization of hardware capabilities, which can translate to faster training or inference and potentially improved model performance. MFU can be computed using either of the following equivalent equations:
\begin{align}\label{eq:1}
    MFU &= \frac{\text{FLOPs/Iter} \times \text{Iter/Sec}}{\text{Device peak FLOPs/Sec}} & MFU &= \frac{\text{Token/Sec} \times \text{FLOPs/Token}}{\text{Device peak FLOPs/Sec}}
\end{align}

Our evaluation results show that our proposed FlowHN framework achieves a much better balance between computational efficiency and downstream task accuracy at every model scale we tested. For the 135M-parameter models, No\_Split attains an average accuracy of 38.85 \%, outperforming the strongest baseline by nearly one full percentage point. It also sustains one of the highest TPS—26443, and also delivers an MFU of 21.99 \%. At the 350M scale, No\_Split again leads on accuracy (41.31 \%), a 1.5 \% improvement over Hymba, while nearly doubling on Hymba’s TPS performance, and also raising MFU to 25.12 \%. Even at the one billion-parameter level, FlowHN variants continue to scale efficiently: FAC\_Split reaches the maximum MFU of 47.63 \%, compared to Hymba’s 33.16 \%, demonstrating that our split based token distribution harnesses hardware capacity better as model size grows, with only a very slight lowering of accuracy.

\begin{table}
  \caption{Comparative analysis of TPS, MFU, and accuracy across various state-of-the-art models, including our proposed FlowHN, evaluated at three model scales: 135M, 350M, and 1B parameters.}
  \label{table:1}
  \centering
  \setlength{\tabcolsep}{3pt}
  \begin{tabular}{ccccccccccc}
    \toprule
    Model Name & Tokens & TPS & MFU & arc\_e & arc\_c & wino & piqa  & hella & boolq & Avg \\
    \midrule
    \multicolumn{11}{c}{\textit{135M}} \\
    \midrule
    Llama   & 1B & 21353 & 22.33 & 27.23 & 22.95 & 47.97 & 52.42 & 25.12 & 41.24 & 36.15 \\
    Mamba-1 & 1B & 11854 & 15.35 & 26.85 & 24.66 & 49.51 & 53.02 & 26.03 & 40.99 & 36.85 \\
    Mamba-2 & 1B & 10094 & 13.29 & 28.16 & 23.89 & 48.54 & 53.13 & 25.45 & 41.08 & 36.71 \\
    Jamba   & 1B & 18324 & 13.83 & 27.78 & 22.44 & 37.81 & 25.79 & 53.02 & 49.35 & 36.03 \\
    SHM     & 1B & 25900 & 22.75 & 29.67 & 22.78 & 40.72 & 26.56 & 55.25 & 50.65 & 37.96 \\
    Hymba   & 1B & 9393  & 12.21 & 29.88 & 23.89 & 49.03 & 55.04 & 25.56 & 44.57 & \underline{37.99} \\
    No\_split\textsubscript{ours} & 1B & 26443 & 21.99 & 29.46 & 24.23 & 49.92 & 56.02 & 27.00 & 43.44 & \textbf{38.85} \\
    AE\_Split\textsubscript{ours} & 1B & 33846 & 22.80 & 26.22 & 23.81 & 48.62 & 52.59 & 25.95 & 40.81 & 36.33 \\
    FA\_Split\textsubscript{ours} & 1B & \underline{34348} & \underline{23.02} & 25.29 & 23.98 & 47.81 & 50.57 & 25.26 & 38.48 & 35.23 \\
    FAC\_Split\textsubscript{ours} & 1B & \textbf{34430} & \textbf{23.07 }& 28.75 & 23.29 & 48.95 & 54.33 & 26.80 & 42.32 & 37.41 \\  
    \midrule
    \multicolumn{11}{c}{\textit{350M}} \\
    \midrule
    Jamba   & 1B & 13202 & 20.03 & 29.04 & 22.61 & 52.14 & 26.71 & 56.78 & 51.22 & 39.75\\
    SHM     & 1B & 8547  & 24.45 & 29.17 & 23.55 & 58.52 & 26.84 & 55.74 & 50.49 & \underline{40.71}\\
    Hymba   & 1B & 4452  & 14.89 & 30.58 & 24.49 & 49.43 & 54.22 & 26.12 & 53.82 & 39.78 \\
    No\_split\textsubscript{ours} & 1B & 8587  & 25.12 & 30.83 & 25.88 & 49.27 & 57.59 & 27.03 & 57.23 & \textbf{41.31} \\
    AE\_Split\textsubscript{ours} & 1B & 13697 & 25.18 & 27.44 & 23.46 & 49.51 & 55.42 & 25.71 & 44.37 & 37.65 \\
    FA\_Split\textsubscript{ours} & 1B & \underline{14246} & \underline{26.41} & 25.42 & 22.53 & 48.54 & 53.08 & 26.01 & 40.81 & 36.07 \\
    FAC\_Split\textsubscript{ours} & 1B & \textbf{14385} & \textbf{26.64} & 29.34 & 25.77 & 47.97 & 56.29 & 27.44 & 56.56 & 40.56 \\  
    \midrule
    \multicolumn{11}{c}{\textit{1B}} \\
    \midrule
    Jamba   & 2B & \textbf{9758} & 40.63 & 34.68 & 23.29 & 49.51 & 58.25 & 26.95 & 56.78 & 41.57\\
    SHM     & 2B & 4923 & 47.35 & 35.56 & 23.89 & 49.76 & 57.49 & 27.44 & 58.25 & 42.06\\
    Hymba   & 2B & 3283 & 33.16 & 37.04 & 21.97 & 49.27 & 59.39 & 27.18 & 60.11 & \underline{42.41}\\
    No\_split\textsubscript{ours} & 2B & 4325 & 45.02 & 36.15 & 22.95 & 49.51 & 59.39 & 28.27 & 61.46 & \textbf{42.95}\\
    AE\_Split\textsubscript{ours} & 2B & 7752 & 46.16 & 33.38 & 22.53 & 49.19 & 55.42 & 26.80 & 53.72 & 40.17 \\
    FA\_Split\textsubscript{ours} & 2B & 7901 & \underline{47.48} & 32.03 & 23.46 & 49.43 & 54.16 & 25.82 & 46.10 & 38.50\\
    FAC\_Split\textsubscript{ours} & 2B & \underline{7926} & \textbf{47.63} & 36.07 & 23.89 & 50.97 & 57.59 & 26.54 & 56.78 & 41.97\\
    \bottomrule
  \end{tabular}
\end{table}

\begin{figure*}[t]
  \centering
   \includegraphics[width=\textwidth]{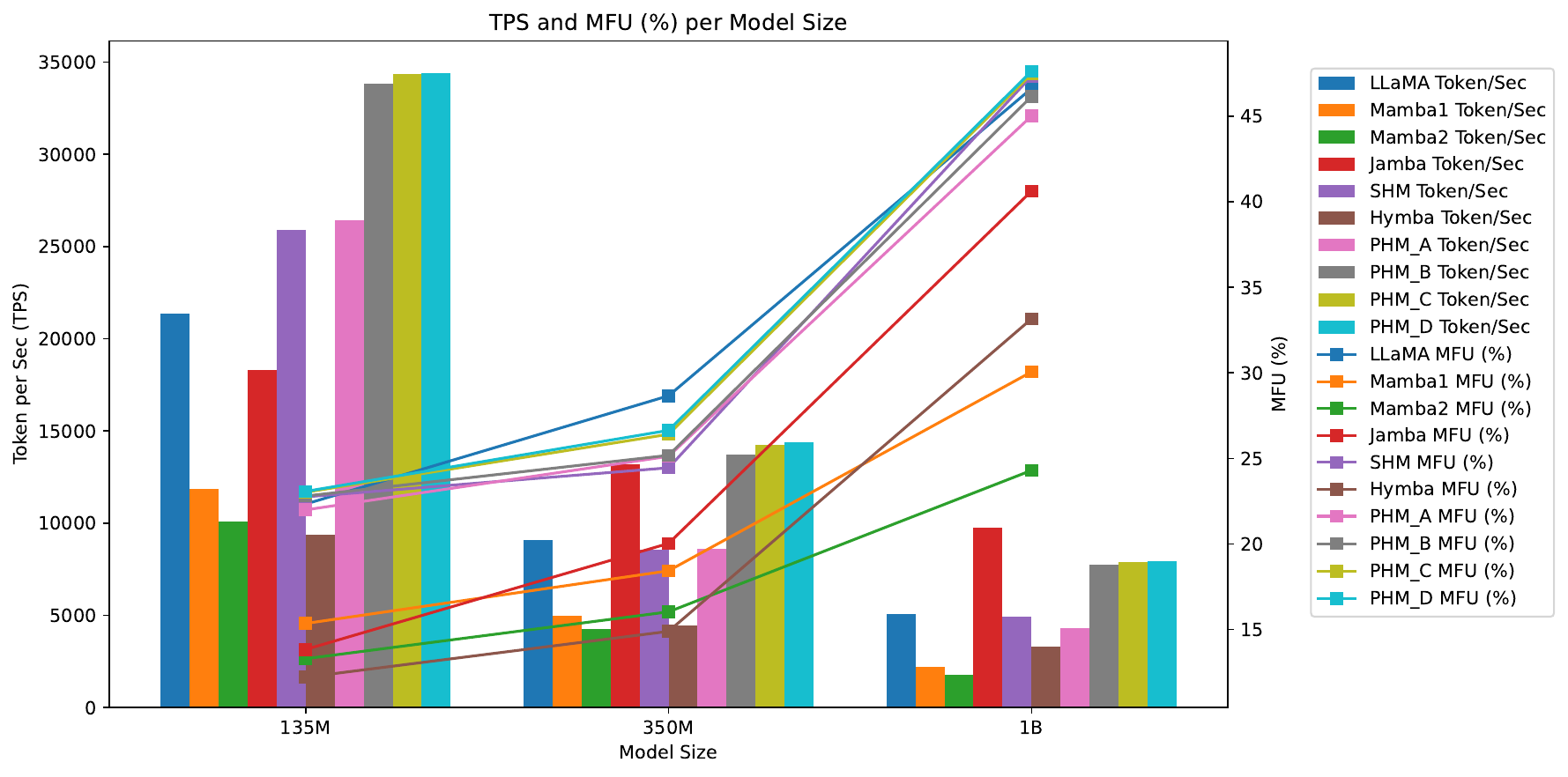}
   \caption{Effect of model scale on TPS and MFU. Bars (left axis) show TPS, while lines (right axis) trace MFU for each model at 135M, 350M, and 1B parameters.}
   \label{fig:3}
\end{figure*}

Figure \ref{fig:3} shows the comparison graph of MFU and TPS between our proposed FlowHN and three state-of-the-art hybrid baselines at three model scales: 135M, 350M, and 1B parameters. Across all sizes, the different token allocation strategies that resort to splitting achieve the highest TPS, with peaks of approximately 34430 TPS at 135M and sustaining over 7926 TPS at 1B. The sole exception is Jamba-1B, which achieves a somewhat higher TPS than FlowHN-1B. This advantage stems from Jamba-1B’s Mixture of Expert (MoE) design: by replacing every other feed-forward block with an MoE layer comprising eight experts, of which only two are active per token, the model maintains a total parameter count of 1B but effectively utilizes only 483M active parameters. The resulting 50\% reduction in active parameters yields faster token processing, at the cost of somewhat lower MFU compared to our FlowHN-1B.

Figure \ref{fig:3} further reveals that FlowHN consistently achieves the highest MFU at all three scales (135M, 350M, and 1B parameters), with one exception: at the 350M scale, LLaMA slightly surpasses FlowHN. This discrepancy arises because LLaMA-350M incurs 27.46T FLOPs/Iter versus FlowHN’s 19.24T, and despite FlowHN’s higher TPS (14385 TPS vs. 9064 TPS), the TPS uplift is insufficient to fully offset the lower compute density—resulting in marginally lower MFU. In contrast, at 135M and 1B scales the FLOPs/Iter gap between LLaMA and FlowHN narrows, so FlowHN’s TPS gains more than compensate, yielding superior MFU across those settings.

Figure \ref{fig:3} also reflects the underlying scaling laws: as model size increases, the total number of floating-point operations per iteration (FLOPs/Iter) grows roughly in proportion to the number of parameters, which drives MFU upward. Conversely, deeper and wider layers incur greater data movement and inter-GPU communication overhead, thereby reducing TPS. According to Equation \ref{eq:1}, since the denominator (theoretical peak FLOPs) remains constant while FLOPs per step increase more than the corresponding drop in Iter/Sec, MFU necessarily improves with scale.

\begin{figure*}[t]
  \centering
   \includegraphics[width=10cm, height=6cm]{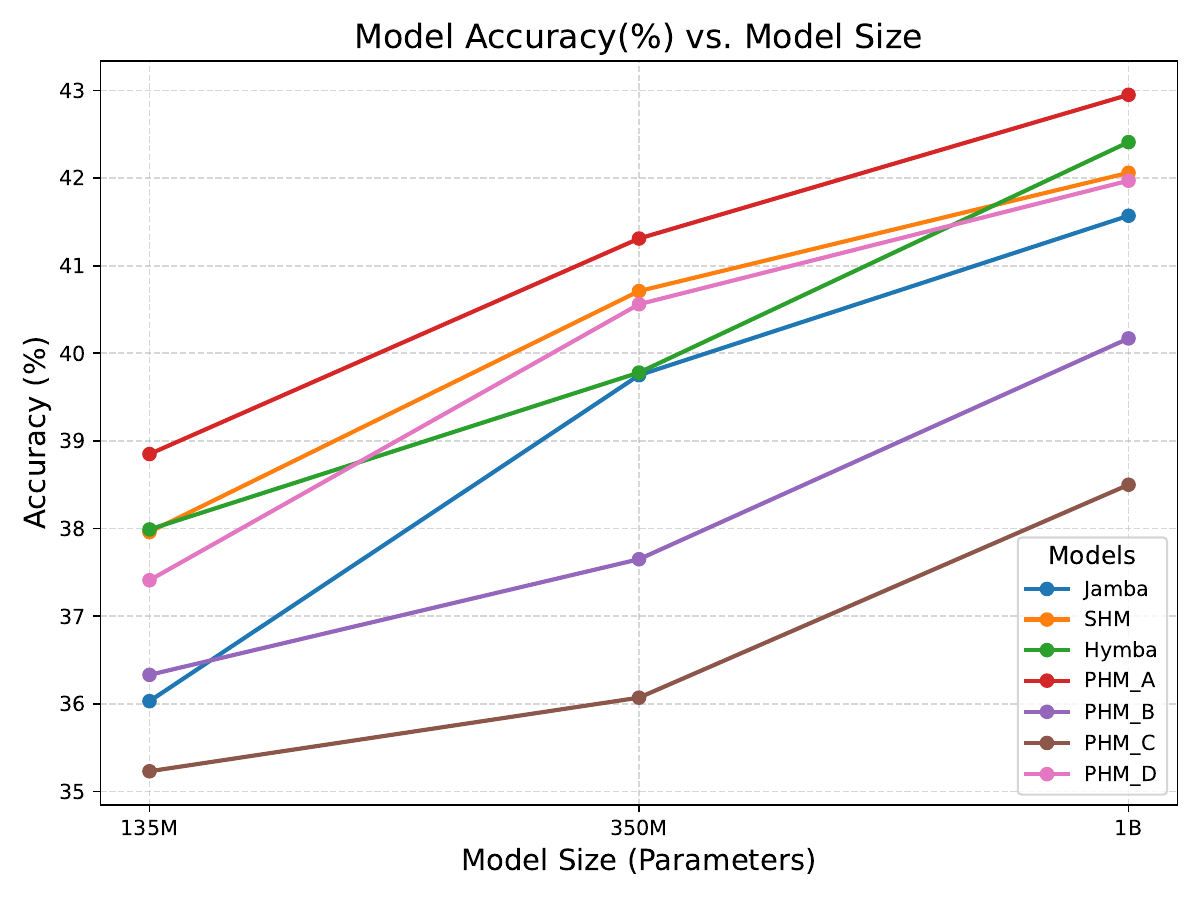}
   \caption{Accuracy as a function of model scale (135 M, 350 M, and 1 B parameters) comparing our approach against state-of-the-art hybrid models.}
   \label{fig:4}
\end{figure*}

Our results in Figure \ref{fig:4} demonstrate the performance of our proposed parallel hybrid model variants relative to state-of-the-art hybrid architectures across three parameter scales. The No\_split mode consistently delivers the largest performance improvements at all scales. FAC\_Split remains competitive with existing hybrids, whereas FA\_Split underperforms on all scales, its token-splitting procedure introduces discontinuities that hinder information flow and yields only modest gains at the 1B parameter scale. FAC\_Split reduces the gap to full-parallel performance, indicating that cyclically reordering tokens so that both Attention and SSM sub-modules observe every token at least once provides benefits, although it still falls short of granting each sub-module complete context.

\section{Conclusion}\label{sec:Conclusion}
In this work, we introduce the FlowHN, a novel architecture that unifies attention and SSMs in a balanced, parallel fashion. By leveraging parallel processing and efficient FLOP aware token distribution, FlowHN ensures that both attention and SSM sub-modules contribute equitably to each block’s workload and addresses the TPS-latency trade-off inherent in traditional hybrid sequence models. Our extensive experiments across multiple model scales demonstrate that FlowHN consistently outperforms existing hybrid parallel architectures, achieves up to $4 \times$ relative gains in Token per Second and delivers up to $2 \times$ gain in MFU. The success of FlowHN underscores the potential of parallel hybrid integration in large-scale AI systems. Future work will explore the scalability of FlowHN to even larger models. We believe that FlowHN paves the way for more efficient and scalable architectures in AI technologies.

\emph{Limitations:} 
The FlowHN has been primarily validated on autoregressive language modeling tasks, such as commonsense reasoning and question answering. Its performance across other NLP tasks, including machine translation, summarization, and recall-intensive applications, remains to be fully explored. Currently, FlowHN employs token-splitting based on computational workload; however, future work could investigate dynamic routing mechanisms that consider both computational efficiency and the relevance or "helpfulness" of each branch for processing specific tokens. Addressing these limitations will be helpful for enhancing the model's robustness and generalization capabilities in our future work.

\emph{Broader Impact:} 
This work proposes a FLOP-aware dynamic token splitting strategy that achieves significant speedups while not losing much on model accuracy. The resulting efficiency improvements have the potential to reduce per-token energy consumption, enable broader access to large-scale models on resource-constrained devices, and facilitate real-time AI applications. However, increased computational efficiency may also accelerate the automation of tasks, potentially requiring re-skilling in certain sectors.

\medskip

\medskip

{
\small

\bibliographystyle{plainnat} 
\bibliography{references}    

}

\end{document}